\documentclass[10pt,twocolumn,letterpaper]{article}

\usepackage[pagenumbers]{cvpr} % To force page numbers, e.g. for an arXiv version
\usepackage{booktabs}
\usepackage{multirow}
\usepackage[table]{xcolor}
\usepackage{graphicx}
\definecolor{cvprblue}{rgb}{0.21,0.49,0.74}
\usepackage[pagebackref,breaklinks,colorlinks,allcolors=cvprblue]{hyperref}

\def\confName{CVPR}
\def\confYear{2027}

\title{Look Closer, Align Better: Explicit Visual Attributes for Robust EEG-to-Image Decoding}

\author{\textbf{Ke Xiao, Banghua Yang, Wenbin Guo, Fenqi Rong}\\[5pt]
School of Mechatronic Engineering
and Automation, Shanghai University, Shanghai, China\\
{\tt\small xiaoke77@shu.edu.cn}
}

\begin{document}
\maketitle
\begin{abstract}
EEG-to-image retrieval commonly aligns neural responses with representations from pretrained vision encoders. However, these representations may underrepresent fine-grained visual details needed to distinguish semantically similar images. To address this limitation, we first investigate the EEG decodability and complementary predictive value of explicit visual attributes, identifying visual cues that capture EEG-relevant information beyond pretrained representations. Building on these findings, we propose \textbf{AttrEcho}, a framework that enriches pretrained visual representations with complementary, spatially aligned visual attributes. Specifically, \textbf{Region-Aligned Visual Attribute Fusion (RAVF)} integrates EEG-relevant local visual attributes with spatially corresponding pretrained features through a learnable signed residual gate. We further employ \textbf{Feature Token Aggregation (FTA)}, which learns dimension-specific weights across regions to aggregate refined visual features for contrastive alignment with EEG representations. Experiments on THINGS-EEG demonstrate that AttrEcho achieves \textbf{94.8\%} Top-1 and \textbf{99.7\%}
Top-5 accuracy in intra-subject 200-way retrieval, while maintaining \textbf{82.1\%} Top-1 accuracy as the gallery expands to 1,600 images with same-concept distractors. AttrEcho also achieves \textbf{25.3\%} Top-1 and \textbf{53.0\%} Top-5 accuracy in inter-subject retrieval. These results highlight the importance of
EEG-relevant visual attributes and spatially aligned fusion for fine-grained EEG-to-image retrieval.

\end{abstract}    
\section{Introduction}
\label{sec:intro}

\begin{figure}[t]
    \centering
    \includegraphics[width=1.0\linewidth]{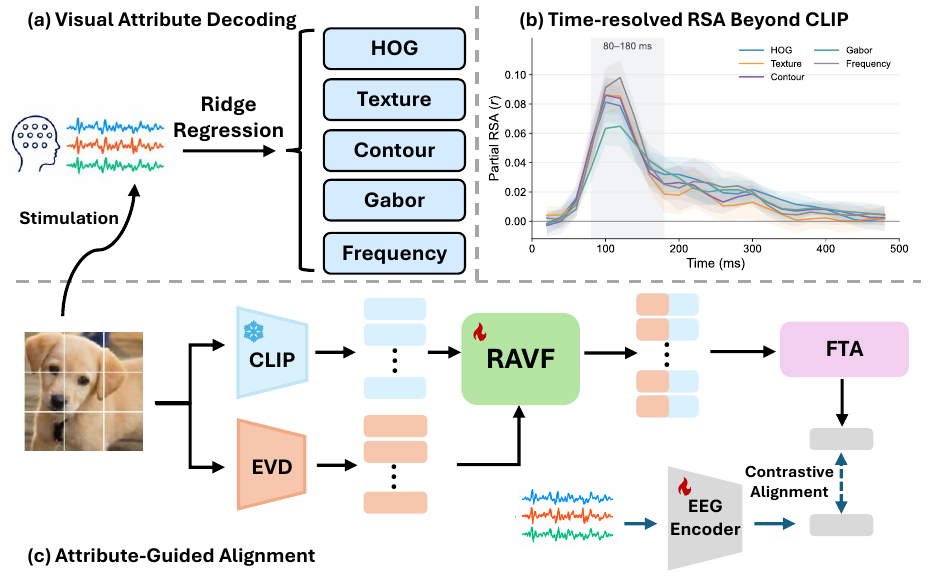}
    \caption{
    Motivation and overview of AttrEcho.
    (a) Visual attribute decoding from EEG signals.
    (b) Time-resolved partial RSA reveals EEG–attribute associations beyond CLIP.
    (c) Overview of the proposed attribute-guided
    EEG--image alignment framework.
    }
    \label{fig:motivation}
\end{figure}

EEG-to-image retrieval aims to identify viewed
images from visually evoked neural responses,
providing a means to investigate how visual
information is represented in the human brain.
Recent methods commonly align EEG features
with representations extracted from pretrained
vision encoders in a shared embedding
space~\cite{song2024decoding,li2024visual,wu2025bridging,liu2026leveraging}.
While advances in EEG representation learning
and cross-modal alignment have substantially
improved retrieval performance, the choice
of visual representations as alignment targets
remains equally important~\cite{wu2025bridging,liu2026leveraging}.
However, it remains unclear whether pretrained
visual representations fully capture the
fine-grained visual attributes reflected
in EEG responses.

Recent studies have enhanced visual alignment targets
through perceptually motivated transformations and
feature selection strategies~\cite{wu2025bridging,liu2026leveraging}. Other approaches explore intermediate visual
features and complementary representations
across the visual hierarchy to improve
brain--vision alignment~\cite{du2026shallow,
tang2026aligning,wang2026visualtarget}. Despite these advances, how explicit, spatially organized visual attributes complement pretrained visual representations remains less understood. Images sharing similar semantic content may still exhibit substantial differences in local visual structure and spatial organization. Such differences may be reflected in EEG responses but are not necessarily emphasized by pretrained visual representations. This motivates us to investigate the relationship between explicit visual attributes and neural responses, aiming to construct more informative alignment targets for EEG-to-image retrieval.

To investigate whether explicit visual attributes
capture EEG-relevant information beyond pretrained
visual representations, we conduct two complementary
analyses, as illustrated in Fig.~\ref{fig:motivation}(a)
and (b). First, we examine whether visual attributes can be decoded from EEG signals using ridge regression, providing evidence that such information is accessible from neural responses. We then perform time-resolved partial representational similarity analysis (RSA)~\cite{kriegeskorte2008rsa} on image pairs within the same object concept, controlling for the representational structure captured by regional CLIP features~\cite{radford2021clip}. The results reveal positive EEG--attribute correspondences beyond those explained by CLIP, particularly during early visual processing. These findings suggest that explicit visual attributes capture aspects of neural representations not fully accounted for by pretrained visual features, highlighting their potential to complement existing visual alignment targets.

Motivated by these findings, we propose \textbf{AttrEcho}, an explicit visual attribute-guided framework for EEG-to-image
retrieval, as illustrated in Fig.~\ref{fig:motivation}(c).
Rather than replacing pretrained visual representations, AttrEcho enriches them with complementary local visual information. Specifically, \textbf{Region-Aligned Visual
Attribute Fusion (RAVF)} projects explicit attributes extracted from corresponding image regions into feature corrections,
which are integrated with regional CLIP features through a learnable signed residual gate. This preserves the original visual features while incorporating spatially aligned attribute information. Furthermore, \textbf{Feature Token
Aggregation (FTA)} learns dimension-specific weights across regions to aggregate the refined visual features into an image-level representation. The resulting representation is projected into a shared embedding space and aligned with EEG features through bidirectional contrastive learning. Our main contributions are summarized as follows:

\begin{itemize}
    \item We investigate the relationship between EEG responses and explicit visual attributes through attribute decoding and within-concept partial RSA, revealing neural representational correspondences not fully explained by pretrained CLIP features.

    \item We propose \textbf{AttrEcho}, an explicit
    visual attribute-guided framework featuring
    \textbf{Region-Aligned Visual Attribute Fusion
    (RAVF)}, which integrates complementary
    local visual attributes with spatially
    corresponding pretrained representations
    through signed gated residual refinement.

    \item Extensive experiments demonstrate the
    effectiveness of AttrEcho, achieving 94.84\%
    Top-1 accuracy in 200-way retrieval and
    82.06\% in 1600-way retrieval with
    same-concept distractors on THINGS-EEG. Further ablations and spatial interventions examine the contributions of complementary visual attributes and region-wise correspondence.

\end{itemize}

\section{Related Work}
\begin{figure*}[t]
    \centering
    \includegraphics[width=0.98\textwidth]{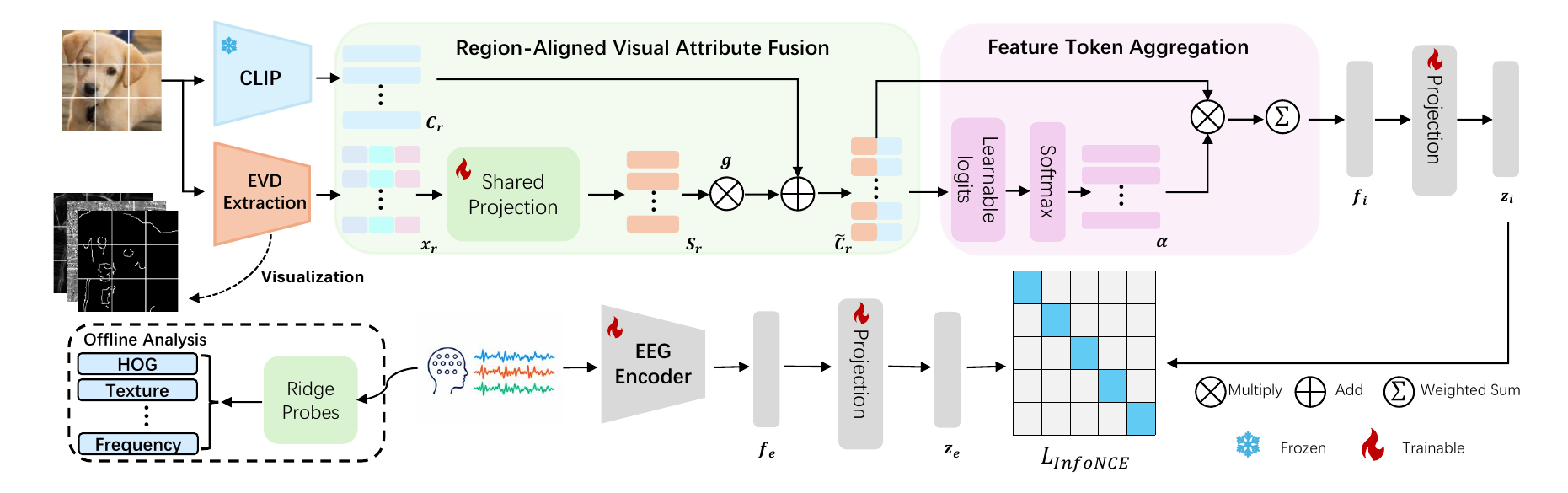}
    \caption{
    Overview of the proposed AttrEcho framework.
    During training, explicit visual descriptors (EVD)
    are integrated with spatially corresponding CLIP features
    through Region-Aligned Visual Attribute Fusion (RAVF).
    The refined features are aggregated using Feature Token
    Aggregation (FTA) and aligned with EEG representations
    using contrastive learning.
    }
    \label{fig:framework}
\end{figure*}

\subsection{EEG-Based Visual Decoding}
Early studies explored joint EEG--visual representations
for image recognition~\cite{palazzo2020decoding}.
NICE established a contrastive learning framework for
zero-shot EEG--image alignment~\cite{song2024decoding}.
Subsequent approaches improved neural representation
learning and cross-modal alignment through enhanced
EEG encoders, cycle consistency, and cognitive
priors~\cite{li2024visual,wei2024mb2c,
zhang2026neurobridge}.
More recently, EEGiT adapted pretrained vision
transformers for EEG encoding~\cite{zhou2026eegit},
while ViEEG introduced hierarchical EEG representations
guided by contour, object, and scene
information~\cite{liu2026vieeg}.
Multimodal approaches further incorporated linguistic
supervision and complementary neural modalities to
enhance visual decoding~\cite{song2025recognizing,
zhang2024cognitioncapturer,xu2026umind,
li2025brainflora}.

Beyond EEG representation learning, increasing attention
has been devoted to the construction of visual alignment
targets. UBP and Blur employed perceptually motivated
blurring strategies to reduce neural--visual
mismatch~\cite{wu2025bridging,liu2026leveraging},
while HyFI integrated perceptual and semantic features
through hyperbolic interpolation~\cite{jo2026hyfi}.
BVSA introduced linguistic priors to decouple
foreground objects from visual
backgrounds~\cite{liu2026linguistic}.
Other studies explored intermediate or hierarchical
visual representations to better match neural
responses~\cite{du2026shallow,zheng2026learning,
tang2026aligning,wang2026visualtarget}.
More closely related to local visual structure,
RAHS investigated correspondences between brain-region
features and visual sub-features~\cite{zhu2026rahs}.
Building on this line of research, our work examines
the EEG decodability of explicit local visual attributes
and their correspondence with EEG responses after
accounting for pretrained CLIP representations.
Rather than replacing or selecting pretrained visual
features, AttrEcho incorporates complementary attributes
as spatially aligned residual refinements.

\subsection{Visual Attributes in Neural Responses}

Neural encoding and decoding studies have long
investigated how visual information is represented
in brain activity. Early fMRI studies demonstrated
that low-level visual features could predict
responses to natural images~\cite{kay2008identifying}.
Subsequent work examined correspondences between
deep visual representations and cortical activity,
as well as the temporal dynamics of object
recognition~\cite{khaligh2014deep,cichy2014resolving}.
Horikawa and Kamitani further demonstrated the
decodability of hierarchical visual features from
fMRI~\cite{horikawa2017generic}, while brain
encoding has been used to examine the hierarchical
organization of pretrained vision
models~\cite{yang2024brain}.

EEG studies have further demonstrated the
relevance of visual attributes to neural responses.
Low-level image statistics have been shown to
confound semantic decoding~\cite{holm2024contribution},
while visual properties such as contrast and
spatial frequency can be decoded from single-trial
EEG responses~\cite{sieghartsleitner2025decoding}.
The decodability of visual attributes has also
been investigated across different EEG electrode
densities~\cite{huang2026robust}.
Time-resolved representational analyses have
revealed distinct neural processing dynamics
associated with object size and depth in natural
images~\cite{lu2025disentangled}.
Recent studies have further examined the roles
of image statistics and spatial structure in
visual perception~\cite{orima2026natural},
as well as the contribution of texture-like
representations to early EEG--DNN
alignment~\cite{loke2026shared}. While previous studies have established
associations between visual attributes and
neural responses, their complementary value
beyond pretrained visual representations
remains insufficiently explored.
Our work systematically examines this
complementarity through neural decoding,
representational analysis, and encoding.

\section{Method}

\subsection{Overview}
Given paired visual stimuli and EEG responses $(I,E)$, our goal is to learn a shared embedding space for EEG-to-image retrieval. As illustrated in Fig.~\ref{fig:framework}, AttrEcho consists of an attribute-guided visual branch and an EEG encoding branch. An offline attribute analysis first evaluates the EEG relevance of candidate
visual descriptors, providing the basis for attribute selection. In the visual branch, each image is divided into nine spatial regions. A frozen CLIP encoder extracts regional visual features, while explicit visual descriptors (EVD) characterize the corresponding local attributes. The proposed Region-Aligned Visual Attribute Fusion (RAVF) module projects these attributes into feature corrections and integrates them with spatially corresponding CLIP features through a learnable signed residual gate. The refined regional features are then aggregated using Feature Token Aggregation (FTA), followed by a projection into the shared embedding space. In parallel, an EEG encoder extracts neural features, which are projected into the same space. The trainable components are optimized through bidirectional contrastive learning to align matched EEG--image pairs.

\subsection{Visual Attribute Analysis}
\begin{figure}[t]
    \centering
    \includegraphics[width=1.0\linewidth]{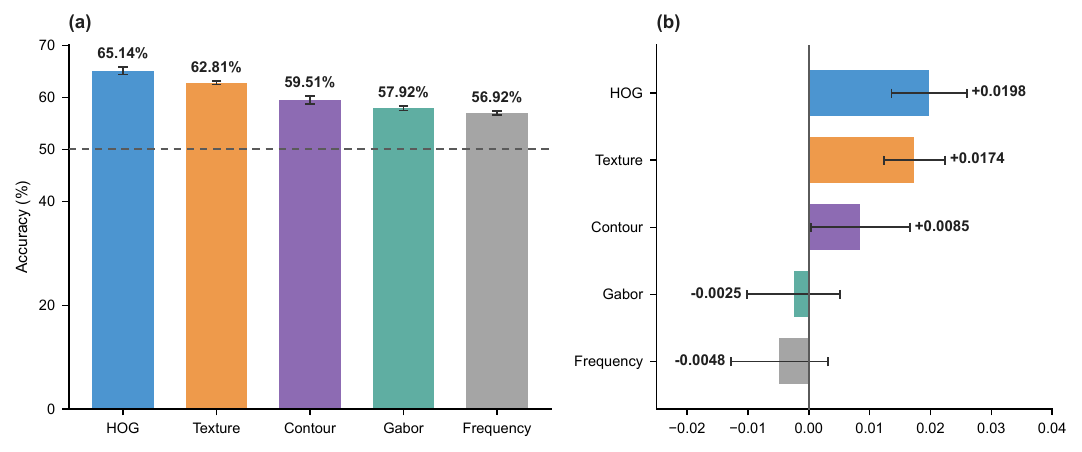}
    \caption{
Visual attribute analysis.
(a) EEG-to-attribute decoding accuracy (2AFC).
(b) Incremental EEG prediction performance beyond CLIP ($\Delta R^2$).
}
    \label{fig:motivation1}
\end{figure}
\label{sec:attribute_analysis}

To examine the EEG relevance and complementary
predictive value of explicit visual attributes,
we conduct decoding and encoding analyses
following established paradigms
~\cite{naselaris2011encoding,horikawa2017generic}.
We consider five visual descriptors:
HOG-based gradients~\cite{dalal2005hog},
LBP- and GLCM-based texture
~\cite{ojala2002lbp,bianconi2021texture},
Canny-based contours~\cite{canny1986edge},
Gabor responses~\cite{zhu2023gabor},
and frequency-domain features.
These descriptors capture complementary
aspects of local image appearance, allowing
us to examine their relationships with EEG
responses and pretrained visual representations.

\noindent\textbf{Time-resolved Partial RSA.}
We perform time-resolved partial RSA~\cite{kriegeskorte2008rsa} to examine
EEG--attribute correspondences beyond
regional CLIP representations.
EEG, attribute, and CLIP features are
standardized using training-set statistics.
Representational dissimilarity matrices
(RDMs) are constructed using cosine distance,
considering only image pairs within the
same object concept.
Let $\mathbf{d}_E(t)$, $\mathbf{d}_A^{(k)}$,
and $\mathbf{d}_C$ denote the vectorized
dissimilarities of the EEG, attribute,
and CLIP RDMs, respectively. We compute the time-resolved partial correlation
$\rho_k(t)$ between EEG responses and visual
attribute $k$ as:

\begin{equation}
\rho_k(t) = \rho_{\mathrm{S}}\left(
\mathbf{d}_E(t), \mathbf{d}_A^{(k)}
\mid \mathbf{d}_C
\right),
\end{equation}
where $\rho_{\mathrm{S}}$ denotes Spearman partial correlation,
and $\mathbf{d}_C$ is used to control for CLIP representational similarity.
As shown in Fig.~\ref{fig:motivation}(b),
positive correlations emerge at early
visual latencies, motivating the
80--180 ms window for subsequent
attribute decoding.

\noindent\textbf{Visual Attribute Decoding.}
We investigate whether explicit visual attributes
can be linearly decoded from EEG responses
~\cite{horikawa2017generic}.
Attribute descriptors from nine image regions
are concatenated, while EEG signals within
the selected time window are flattened and used to train
attribute-specific ridge regressors:
\begin{equation}
    \hat{\mathbf{a}}^{(k)}
    = \mathbf{e}W_k + \mathbf{b}_k,
    \label{eq:attribute_decoding}
\end{equation}
where $\mathbf{e}$ denotes EEG features and
$k$ indexes the attribute.
The training data are split into concept-disjoint
training, validation, and held-out sets, with
regularization selected on the validation set.

Decodability is evaluated using within-concept
two-alternative forced-choice (2AFC) identification.
Predicted attributes are compared with candidate descriptors using cosine similarity. Each target is paired with every
same-concept distractor, and accuracy is averaged
over all pairs. As shown in
Fig.~\ref{fig:motivation1}(a), all five attributes
exceed the 50\% chance level, with HOG, Texture,
and Contour achieving the highest accuracies.

\noindent\textbf{Conditional EEG Encoding.}
We further investigate whether explicit visual
attributes provide additional predictive information
beyond pretrained CLIP representations.
Following regularized neural encoding approaches
\cite{dupre2022featurespace,lin2024stacked},
we compare two ridge encoding models that predict
EEG responses using either CLIP features alone
or CLIP features augmented with an individual
visual attribute.
The incremental predictive contribution is measured as:
% \begin{equation}
%     \Delta R_k^2 =
%     R^2(\mathrm{CLIP}+\mathbf{a}^{(k)})
%     - R^2(\mathrm{CLIP}),
%     \label{eq:conditional_encoding}
% \end{equation}

\begin{equation}
\Delta R_k^2 =
R^2([\mathbf{z}_C;\mathbf{z}_A^{(k)}])
- R^2(\mathbf{z}_C),
\end{equation}

where $\mathbf{z}_C$ and $\mathbf{z}_A^{(k)}$ denote
the CLIP and attribute features, respectively,
and $R^2$ is the coefficient of determination for EEG
prediction. Positive values indicate improved prediction
after incorporating the attribute.
As shown in Fig.~\ref{fig:motivation1}(b),
HOG, Texture, and Contour yield positive
incremental contributions, whereas Gabor and
Frequency do not improve prediction under
the evaluated setting.

Together, these analyses demonstrate the
decodability of explicit visual attributes
from EEG and identify attributes that provide
complementary predictive information beyond CLIP. These findings motivate the use of HOG, Texture,
and Contour for region-aligned visual refinement
in AttrEcho.

\subsection{Region-Aligned Visual Attribute Fusion}
\label{sec:ravf}

To incorporate EEG-relevant visual attributes into
pretrained image representations, we introduce
Region-Aligned Visual Attribute Fusion (RAVF),
which performs attribute-guided refinement at
the regional level. Given an input image, we
extract nine spatially organized regions and obtain their visual embeddings using a frozen CLIP-RN50 encoder. For each region, we extract three
explicit visual descriptors: HOG, Texture,
and Contour. These descriptors are concatenated
to form a regional attribute representation:
\begin{equation}
    \mathbf{a}_r =
    [\mathbf{h}_r;\mathbf{t}_r;\mathbf{o}_r]
    \in \mathbb{R}^{d_a},
    \label{eq:attribute_representation}
\end{equation}
where $r$ indexes the image region, and
$\mathbf{h}_r$, $\mathbf{t}_r$, and $\mathbf{o}_r$
denote its HOG, Texture, and Contour
descriptors, respectively, and $d_a$ is
the dimension of the concatenated attribute vector.

To align explicit attributes with pretrained
visual representations, a shared projection
network maps each regional attribute vector
into the CLIP feature space. The projector consists of layer normalization, two linear layers with an intermediate ELU activation, and a Tanh output activation:

\begin{equation}
    \mathbf{s}_r =
    \tanh\left(
    \operatorname{MLP}
    \left(\operatorname{LN}(\mathbf{a}_r)\right)
    \right),
    \label{eq:attribute_projection}
\end{equation}
where the projection network shares its parameters across all nine regions. Rather than replacing pretrained visual features,
RAVF incorporates the projected attributes through
gated residual refinement. Specifically, each regional attribute representation is fused exclusively with the CLIP feature from the same spatial location:
\begin{equation}
    \tilde{\mathbf{c}}_r
    = \mathbf{c}_r + g\mathbf{s}_r,
    \label{eq:ravf}
\end{equation}
where $\mathbf{c}_r$ denotes the frozen CLIP
feature and $g$ is a bounded, signed gate controlled by a learnable
scalar, initialized to zero and
shared across all regions. This formulation
preserves the original CLIP features while
allowing attribute-based residual corrections
during training. 

% The refined regional features
% are then aggregated into an image-level
% representation.

\subsection{Feature Aggregation and Alignment}
\label{sec:alignment}
To construct an image-level representation,
we employ Feature Token Aggregation (FTA)
to aggregate the nine refined regional features.
Feature-wise weighted pooling has been explored
in previous studies~\cite{sun2017learning,
yang2020attsets}, including related feature
selection strategies for EEG-based visual
decoding~\cite{liu2026leveraging}.
Here, we learn dimension-specific aggregation
weights across the nine spatial regions. Specifically, given the refined regional
features $\mathbf{T}\in\mathbb{R}^{9\times D}$,
the aggregation weights are computed as:
\begin{equation}
    \alpha_{r,d} =
    \frac{\exp(A_{r,d})}
    {\sum_{j=1}^{9}\exp(A_{j,d})},
    \label{eq:fta_weight}
\end{equation}
where $\mathbf{A}\in\mathbb{R}^{9\times D}$
contains learnable aggregation logits.
The image-level representation is obtained by
\begin{equation}
    f_{i,d} = \sum_{r=1}^{9}
    \alpha_{r,d}T_{r,d},
    \label{eq:fta_aggregation}
\end{equation}

The aggregated feature vector is subsequently
passed through an image projection head to
obtain the visual embedding $\mathbf{z}_i$. For the neural branch, we adopt an EEG encoder
that combines spatial convolution and temporal
feature transformation to extract representations
from stimulus-evoked EEG signals. 
An EEG projection head maps the encoded features
into the shared embedding space, producing $\mathbf{z}_e$. Both embeddings are $\ell_2$-normalized before cross-modal alignment.

Given a batch of $N$ matched EEG--image pairs,
we optimize the bidirectional contrastive loss:
\begin{equation}
\begin{aligned}
\mathcal{L} = -\frac{1}{2N}\sum_{n=1}^{N}
\Bigg[
&\log
\frac{\exp\left(
\mathbf{z}_e^{n}
\mathbf{z}_i^{n}/\tau\right)}
{\sum_{j=1}^{N}\exp\left(
\mathbf{z}_e^{n}
\mathbf{z}_i^{j}/\tau\right)}
\\
+&\log
\frac{\exp\left(
\mathbf{z}_i^{n}
\mathbf{z}_e^{n}/\tau\right)}
{\sum_{j=1}^{N}\exp\left(
\mathbf{z}_i^{n}
\mathbf{z}_e^{j}/\tau\right)}
\Bigg],
\end{aligned}
\label{eq:contrastive}
\end{equation}
where $\tau$ is the temperature parameter. During inference, each candidate image is
encoded by the visual branch, while the
averaged EEG response is mapped into the
shared embedding space. Candidate images
are ranked according to their cosine
similarity with the EEG query, and the
highest-ranked image is returned as the
retrieval result.

\section{Experiments}
\label{sec:experiments}
\subsection{Experimental Setup}
\label{sec:experimental_setup}
\noindent\textbf{Datasets.}
We evaluate our model on three publicly available
datasets. \textbf{THINGS-EEG}~\cite{gifford2022large} contains EEG recordings from 10 subjects using 63 channels, with 16,540 training images from 1,654 concepts and 200 held-out test images. \textbf{THINGS-MEG}~\cite{hebart2023things} contains MEG recordings from four subjects using 271 channels, covering 22,248 natural images
and 200 test images repeated 12 times. \textbf{Alljoined-1.6M}~\cite{xu2025alljoined} comprises over 1.6 million EEG trials from 20 subjects, recorded using a 32-channel consumer-grade EEG system. It shares the same 16,540 training and 200 test images as THINGS-EEG.

\noindent\textbf{Baseline Methods.} We compare AttrEcho with eight representative EEG-to-image retrieval methods. BraVL \cite{du2023decoding} integrates brain, visual, and linguistic representations through multimodal learning. NICE \cite{song2024decoding} establishes EEG–image alignment using contrastive learning, while ATM \cite{li2024visual} employs an attention-based EEG encoder for neural representation learning. UBP \cite{wu2025bridging} introduces uncertainty-aware image blurring to alleviate visual–neural discrepancies. BVSA \cite{liu2026linguistic} leverages linguistic priors to decouple foreground objects from backgrounds, while NeuroBridge \cite{zhang2026neurobridge} incorporates cognitive prior augmentation and bidirectional semantic alignment. Blur \cite{liu2026leveraging} exploits multi-level blurred visual features with learned feature selection, whereas Shallow Alignment \cite{du2026shallow} improves alignment by leveraging intermediate visual representations. 
% These methods provide comprehensive comparisons across neural encoding, multimodal alignment, and visual representation refinement.

\noindent\textbf{Evaluation Protocol.}
We evaluate EEG-to-image retrieval under both intra-subject and inter-subject settings. For intra-subject evaluation, a separate model is trained and evaluated for each subject, while inter-subject evaluation follows a leave-one-subject-out protocol. Each EEG query is matched against candidate images using cosine similarity between learned embeddings. We report Top-1 and Top-5 accuracy under the 200-way zero-shot retrieval setting. To assess retrieval performance with larger candidate spaces, we expand the gallery to 400, 800, and 1,600 images by introducing additional distractors. For large-gallery comparisons, identical candidate sets are used across methods.

\noindent\textbf{Implementation Details.}
We optimize all trainable components using AdamW with an initial learning rate of $1\times10^{-3}$ and a weight decay of $0.01$. The EEG and image projection heads consist of a linear layer and a two-layer MLP, respectively, both producing 1,024-dimensional embeddings. Models are trained for up to 200 epochs with a batch size of 1,024, without a learning rate scheduler. The checkpoint with the highest validation Top-1 accuracy is selected for evaluation. Each test was repeated five times to ensure stability of the results. All experiments are conducted on an NVIDIA GeForce RTX 4090 GPU.

\subsection{Comparison with SOTA Methods}

\begin{table*}[t]
\centering
\caption{Top-1 (T1) and Top-5 (T5) accuracy (\%) for 200-way zero-shot retrieval on THINGS-EEG.}
\label{tab:main_results}

\setlength{\tabcolsep}{3.0pt}
\renewcommand{\arraystretch}{1.05}

\resizebox{\textwidth}{!}{
\begin{tabular}{l c *{11}{cc}}
\toprule

\multirow{2}{*}{\textbf{Method}} &
\multirow{2}{*}{\textbf{Year}} &
\multicolumn{2}{c}{Subject 1} &
\multicolumn{2}{c}{Subject 2} &
\multicolumn{2}{c}{Subject 3} &
\multicolumn{2}{c}{Subject 4} &
\multicolumn{2}{c}{Subject 5} &
\multicolumn{2}{c}{Subject 6} &
\multicolumn{2}{c}{Subject 7} &
\multicolumn{2}{c}{Subject 8} &
\multicolumn{2}{c}{Subject 9} &
\multicolumn{2}{c}{Subject 10} &
\multicolumn{2}{c}{Avg} \\

\cmidrule(lr){3-4}
\cmidrule(lr){5-6}
\cmidrule(lr){7-8}
\cmidrule(lr){9-10}
\cmidrule(lr){11-12}
\cmidrule(lr){13-14}
\cmidrule(lr){15-16}
\cmidrule(lr){17-18}
\cmidrule(lr){19-20}
\cmidrule(lr){21-22}
\cmidrule(lr){23-24}

& &
T1 & T5 &
T1 & T5 &
T1 & T5 &
T1 & T5 &
T1 & T5 &
T1 & T5 &
T1 & T5 &
T1 & T5 &
T1 & T5 &
T1 & T5 &
T1 & T5 \\
\midrule

\multicolumn{24}{c}{\textbf{Intra-subject: train and test on one subject}} \\
\midrule

BraVL~\cite{du2023decoding} &
2023 &
6.1 & 17.9 &
4.9 & 14.9 &
5.6 & 17.4 &
5.0 & 15.1 &
4.0 & 13.4 &
6.0 & 18.2 &
6.5 & 20.4 &
8.8 & 23.7 &
4.3 & 14.0 &
7.0 & 19.7 &
5.8 & 17.5 \\

NICE~\cite{song2024decoding} &
2024 &
13.2 & 39.5 &
13.5 & 40.3 &
14.5 & 42.7 &
20.6 & 52.7 &
10.1 & 31.5 &
16.5 & 44.0 &
17.0 & 42.1 &
22.9 & 56.1 &
15.4 & 41.6 &
17.4 & 45.8 &
16.1 & 43.6 \\

ATM~\cite{li2024visual} &
2024 &
25.6 & 60.4 &
22.0 & 54.5 &
25.0 & 62.4 &
31.4 & 60.9 &
12.9 & 43.0 &
21.3 & 51.1 &
30.5 & 61.5 &
38.8 & 72.0 &
34.4 & 51.5 &
29.1 & 63.5 &
27.1 & 58.1 \\
UBP~\cite{wu2025bridging} &
2025 &
41.2 & 70.5 &
51.2 & 80.9 &
51.2 & 82.0 &
51.1 & 76.9 &
42.2 & 72.8 &
57.5 & 83.5 &
49.0 & 79.9 &
58.6 & 85.8 &
45.1 & 76.2 &
61.5 & 88.2 &
50.9 & 79.7 \\

BVSA~\cite{liu2026linguistic} &
2026 &
57.0 & 85.5 &
55.5 & 89.0 &
57.5 & 92.0 &
54.5 & 88.5 &
51.0 & 81.5 &
61.5 & 92.0 &
56.5 & 85.5 &
66.0 & 92.5 &
55.0 & 89.0 &
67.0 & 95.5 &
58.2 & 89.1 \\

NeuroBridge~\cite{zhang2026neurobridge} &
2026 &
50.0 & 77.6 &
63.2 & 90.6 &
61.6 & 91.1 &
61.4 & 90.0 &
54.8 & 85.0 &
69.7 & 92.9 &
62.7 & 88.8 &
71.2 & 95.1 &
64.0 & 91.0 &
73.6 & 97.1 &
63.2 & 89.9 \\

% EEGiT~\cite{Zhou_2026_CVPR} &
% 2026 &
% 82.3 & 97.6 &
% 73.7 & 97.9 &
% 67.5 & 95.7 &
% 64.9 & 93.5 &
% 55.9 & 91.3 &
% 66.8 & 92.0 &
% 70.2 & 98.1 &
% 84.8 & 98.6 &
% 51.7 & 87.3 &
% 86.5 & 99.4 &
% 70.4 & 95.1 \\

% BrainHIVE~\cite{zheng2026learning} &
% 2026 &
% 64.3 & 88.8 &
% 76.3 & 95.3 &
% 74.0 & 95.0 &
% 67.0 & 91.8 &
% 68.0 & 91.5 &
% 81.5 & 96.3 &
% 76.8 & 96.8 &
% 84.8 & 98.5 &
% 76.8 & 95.8 &
% 87.3 & 99.3 &
% 75.7 & 94.6 \\

Blur~\cite{liu2026leveraging} &
2026 &
81.9 & 96.8 &
81.7 & 96.8 &
78.3 & 96.0 &
76.9 & 97.2 &
71.4 & 93.7 &
84.3 & 98.7 &
78.2 & 96.7 &
84.5 & 98.1 &
78.7 & 96.8 &
84.1 & 98.1 &
80.0 & 96.9 \\

Shallow ~\cite{du2026shallow} &
2026 &
75.0 & 94.3 &
87.5 & 98.9 &
83.2 & 98.2 &
79.5 & 96.1 &
74.6 & 96.4 &
89.9 & 99.3 &
78.5 & 97.3 &
86.9 & 99.4 &
81.3 & 97.8 &
89.3 & 99.1 &
82.6 & 97.7 \\
\rowcolor{green!20}
\textbf{AttrEcho(Ours)} &
2026 &
\textbf{95.9} & \textbf{99.6} &
\textbf{96.1} & \textbf{99.9} &
\textbf{93.7} & \textbf{99.9} &
\textbf{90.1} & \textbf{99.5} &
\textbf{93.6} & \textbf{99.0} &
\textbf{96.1} & \textbf{99.6} &
\textbf{95.5} & \textbf{99.9} &
\textbf{96.7} & \textbf{100.0} &
\textbf{93.0} & \textbf{99.3} &
\textbf{97.7} & \textbf{99.9} &
\textbf{94.8} & \textbf{99.7} \\

\midrule

\multicolumn{24}{c}{\textbf{Inter-subject: leave one subject out for test}} \\
\midrule

BraVL~\cite{du2023decoding} &
2023 &
2.3 & 8.0 &
1.5 & 6.3 &
1.4 & 5.9 &
1.7 & 6.7 &
1.5 & 5.6 &
1.8 & 7.2 &
2.1 & 8.1 &
2.2 & 7.6 &
1.6 & 6.4 &
2.3 & 8.5 &
1.8 & 7.0 \\
NICE~\cite{song2024decoding} &
2024 &
7.6 & 22.8 &
5.9 & 20.5 &
6.0 & 22.3 &
6.3 & 20.7 &
4.4 & 18.3 &
5.6 & 22.2 &
5.6 & 19.7 &
6.3 & 22.0 &
5.7 & 17.6 &
8.4 & 28.3 &
6.2 & 21.4 \\

ATM~\cite{li2024visual} &
2024 &
10.5 & 26.8 &
7.1 & 24.8 &
11.9 & 33.8 &
14.7 & 39.4 &
7.0 & 23.9 &
11.1 & 35.8 &
16.1 & 43.5 &
15.0 & 40.3 &
4.9 & 22.7 &
20.5 & 46.5 &
11.9 & 33.8 \\

UBP~\cite{wu2025bridging} &
2025 &
11.5 & 29.7 &
15.5 & 40.0 &
9.8 & 27.0 &
13.0 & 32.3 &
8.8 & 33.8 &
11.7 & 31.0 &
10.2 & 23.8 &
12.2 & 32.2 &
15.5 & 40.5 &
16.0 & 43.5 &
12.4 & 33.4 \\

% D$^{2}$-FOSA~\cite{ubp} &
% 2026 &
% 12.5 & 33.5 &
% 13.5 & 37.5 &
% 11.5 & 29.5 &
% 13.5 & 33.5 &
% 11.5 & 34.5 &
% 12.0 & 33.0 &
% 12.5 & 32.5 &
% 12.5 & 32.5 &
% 13.5 & 35.5 &
% 17.5 & 43.5 &
% 13.1 & 34.6 \\

BVSA~\cite{liu2026linguistic} &
2026 &
17.5 & 41.5 &
21.0 & 46.0 &
7.5 & 27.5 &
17.0 & 41.0 &
13.0 & 33.5 &
16.0 & 36.0 &
14.0 & 35.5 &
13.0 & 39.0 &
18.0 & 43.5 &
21.5 & 50.0 &
15.8 & 39.4 \\

NeuroBridge~\cite{zhang2026neurobridge} &
2026 &
23.2 & 52.4 &
21.2 & 49.3 &
13.2 & 36.5 &
17.0 & 45.3 &
14.5 & 37.7 &
\cellcolor{green!20}\textbf{25.0} & \cellcolor{green!20}\textbf{55.0} &
15.3 & 45.1 &
20.1 & 44.9 &
13.7 & 36.5 &
27.2 & 56.3 &
19.0 & 45.9 \\

% EEGiT~\cite{Zhou_2026_CVPR} &
% 2026 &
% 26.9 & 57.2 &
% 22.9 & 51.4 &
% 14.5 & 41.8 &
% 20.5 & 50.2 &
% 27.1 & 56.4 &
% 26.0 & 63.4 &
% 22.0 & 48.5 &
% 20.2 & 50.7 &
% 26.5 & 67.3 &
% 33.7 & 69.4 &
% 24.0 & 55.6 \\

Blur~\cite{liu2026leveraging} &
2026 &
27.3 & 54.9 &
32.0 & \cellcolor{green!20}\textbf{63.0} &
11.1 & 34.5 &
18.6 & 46.0 &
16.9 & 40.7 &
16.1 & 43.2 &
16.0 & 45.4 &
18.3 & 49.0 &
13.6 & 41.2 &
30.1 & 62.1 &
20.0 & 48.0 \\
% MB$^{2}$L~\cite{mb2l} &
% 2026 &
% 22.5 & 62.5 &
% 30.0 & 57.0 &
% 16.5 & 35.5 &
% 20.5 & 43.5 &
% 12.0 & 32.0 &
% 16.5 & 41.5 &
% 21.0 & 43.5 &
% 15.5 & 35.0 &
% 17.0 & 42.0 &
% 26.5 & 54.5 &
% 19.8 & 44.7 \\

Shallow~\cite{du2026shallow} &
2026 &
23.5 & 53.2 &
30.6 & 60.4 &
10.0 & 28.1 &
19.5 & \cellcolor{green!20}\textbf{48.3} &
18.1 & \cellcolor{green!20}\textbf{45.2} &
22.7 & 49.8 &
18.6 & 46.0 &
17.3 & 46.1 &

\cellcolor{green!20}\textbf{23.0} & \cellcolor{green!20}\textbf{54.8} &
34.4 & 62.0 &
21.8 & 49.4 \\

\rowcolor{green!20}
\textbf{AttrEcho(Ours)} &
2026 &
\textbf{32.4} & \textbf{68.2} &
\textbf{33.2} & \textbf{62.2} &
\textbf{17.0} & \textbf{39.8} &
\textbf{20.6} & \textbf{44.9} &
\textbf{19.3} & \textbf{44.9} &
\textbf{22.1} & \textbf{50.0} &
\textbf{28.5} & \textbf{54.5} &
\textbf{24.1} & \textbf{53.2} &
\textbf{19.4} & \textbf{44.9} &
\textbf{36.7} & \textbf{67.6} &
\textbf{25.3} & \textbf{53.0} \\

\bottomrule
\end{tabular}
}
\end{table*}

We compare AttrEcho with eight EEG-to-image retrieval methods on THINGS-EEG under both intra-subject and inter-subject settings. As shown in Table~\ref{tab:main_results}, our method consistently outperforms existing approaches in the intra-subject setting, achieving average Top-1 and Top-5 accuracies of 94.8\% and 99.7\%, respectively. Compared with the strongest baseline, these results represent improvements of 12.2\% and 2.0\%, with the best performance achieved across all ten subjects. In the more challenging inter-subject setting, our method achieves average Top-1 and Top-5 accuracies of 25.3\% and 53.0\%, surpassing Shallow by 3.5\% and 3.6\%, respectively. Notably, our method achieves the highest Top-1 accuracy on eight out of ten held-out subjects.

We further evaluate our method on THINGS-MEG and Alljoined-1.6M to assess its effectiveness across different datasets. As shown in Table~\ref{tab:multi_dataset}, our method achieves Top-1 and Top-5 accuracies of 59.3\% and 82.2\% on THINGS-MEG under the intra-subject setting, outperforming the strongest baseline, Shallow, by 11.3\% and 7.8\%, respectively. Under the inter-subject setting, our method also achieves the best performance, with Top-1 and Top-5 accuracies of 6.7\% and 18.9\%. On Alljoined-1.6M, our method achieves 22.5\% Top-1 and 48.8\% Top-5 accuracy in the intra-subject setting, exceeding Blur by 12.1\% and 18.6\%, respectively. Furthermore, it achieves 9.3\% Top-1 and 27.1\% Top-5 accuracy under the inter-subject setting. These results further demonstrate the effectiveness of our method across different neural decoding datasets.

\begin{table}[t]
\centering
\caption{Comparison with state-of-the-art methods on THINGS-MEG and Alljoined-1.6M.}
\label{tab:multi_dataset}
\renewcommand{\arraystretch}{1.5}
\setlength{\tabcolsep}{4pt}

\resizebox{\columnwidth}{!}{
\begin{tabular}{lcccccccc}
\toprule[1.2pt]

\multirow{3}{*}{\textbf{Method}}
& \multicolumn{4}{c}{\textbf{THINGS-MEG}}
& \multicolumn{4}{c}{\textbf{Alljoined-1.6M}} \\

\cmidrule(lr){2-5}
\cmidrule(lr){6-9}

& \multicolumn{2}{c}{\textbf{Intra-subject}}
& \multicolumn{2}{c}{\textbf{Inter-subject}}
& \multicolumn{2}{c}{\textbf{Intra-subject}}
& \multicolumn{2}{c}{\textbf{Inter-subject}} \\

& Top-1 & Top-5
& Top-1 & Top-5
& Top-1 & Top-5
& Top-1 & Top-5\\

\midrule

UBP~\cite{wu2025bridging}
& 26.7 & 55.2 & 2.2 & 10.4
& 8.5 & 25.1 & 4.2 & 15.0 \\

NeuroBridge~\cite{zhang2026neurobridge}
& 32.2 & 60.8 & 3.4 & 12.8
& 8.7 & 26.4 & 4.8 & 17.5 \\

Blur~\cite{liu2026leveraging} 
& 44.0 & 72.0 & 5.3 & 15.9
& 10.4 & 30.2 & 8.6 & 25.7 \\

Shallow~\cite{du2026shallow}
& 48.0 & 74.4 & 3.7 & 13.0
& 10.1 & 28.4 & 6.5 & 21.4 \\

\midrule
\rowcolor{green!20}
\textbf{AttrEcho}
& \textbf{59.3} & \textbf{82.2}
& \textbf{6.7} & \textbf{18.9}
& \textbf{22.5} & \textbf{48.8} & \textbf{9.3} & \textbf{27.1} \\

\bottomrule[1.2pt]
\end{tabular}
}
\end{table}

\subsection{Large-Gallery Retrieval}
To evaluate retrieval robustness as gallery size
increases, we expand the THINGS-EEG gallery from
200 to 400, 800, and 1,600 images.
Each gallery retains the 200 official test images
and adds 1, 3, or 7 same-concept distractors per
concept. Only the exact viewed image is considered correct. For each enlarged gallery size, we independently
sample 20 gallery configurations without
replacement within each concept.
All methods use identical galleries and candidate
ordering, with fixed model checkpoints and no
additional fine-tuning or checkpoint selection.
Results are averaged across gallery configurations,
model runs, and ten subjects.

As shown in Table~\ref{tab:large_gallery}, our
method consistently outperforms existing approaches
across all gallery sizes. At 1,600-way retrieval,
our method achieves 82.1\% Top-1 and 96.6\%
Top-5 accuracy, surpassing the strongest baseline,
Shallow, by 25.7\% and 12.8\%,
respectively. Notably, our method exhibits a
substantially smaller Top-1 degradation of only
12.7\% from 200-way to 1,600-way,
compared with 26.2\% for Shallow and 29.1\% for Blur. Moreover, the Top-1 advantage over Shallow increases from 12.2\% to 25.7\% as the gallery
expands. These results demonstrate the robustness
of our method against increasing same-concept
distractors and its effectiveness in fine-grained
instance-level retrieval.

\begin{table}[t]
\centering
\caption{Retrieval performance under increasing
gallery sizes on THINGS-EEG (\%).
Drop denotes the absolute decrease in Top-1 accuracy
from 200-way to 1600-way retrieval. Performance gains are measured against Shallow.}
% Improvements over Shallow are reported in
% percentage points (pp).}
\label{tab:large_gallery}

\setlength{\tabcolsep}{3pt}
\renewcommand{\arraystretch}{1.5}

\resizebox{\columnwidth}{!}{%
\begin{tabular}{lccccccccc}
\toprule[1.1pt]
\multirow{2}{*}{\textbf{Method}}
& \multicolumn{2}{c}{\textbf{200-way}}
& \multicolumn{2}{c}{\textbf{400-way}}
& \multicolumn{2}{c}{\textbf{800-way}}
& \multicolumn{2}{c}{\textbf{1600-way}}
& \multirow{2}{*}{\textbf{Drop $\downarrow$}} \\
\cmidrule(lr){2-3}
\cmidrule(lr){4-5}
\cmidrule(lr){6-7}
\cmidrule(lr){8-9}
& Top-1 & Top-5
& Top-1 & Top-5
& Top-1 & Top-5
& Top-1 & Top-5 & \\
\midrule

NeuroBridge~\cite{zhang2026neurobridge}
& 63.2 & 89.9
& 51.1 & 82.0
& 41.2 & 72.1
& 32.3 & 60.9
& 30.9 \\

Blur~\cite{liu2026leveraging}
& 80.0 & 96.9
& 70.7 & 93.3
& 61.4 & 87.8
& 50.9 & 80.7
& 29.1 \\

Shallow~\cite{du2026shallow}
& 82.6 & 97.7
& 74.1 & 94.8
& 65.6 & 90.2
& 56.4 & 83.8
& 26.2 \\

\midrule
\rowcolor{green!20}
& \textbf{94.8} & \textbf{99.7}
& \textbf{91.8} & \textbf{99.3}
& \textbf{87.5} & \textbf{98.2}
& \textbf{82.1} & \textbf{96.6}
& \textbf{12.7} \\

\rowcolor{green!20}
\multirow{-2}{*}{\textbf{AttrEcho}}
& (+12.2) &(+2.0) 
&(+17.7)  &(+4.5) 
&(+21.9)  &(+8.0) 
& (+25.7) & (+12.8) 
& (-13.5)  \\

\bottomrule[1.1pt]
\end{tabular}%
}
\end{table}

\subsection{Ablation Studies}
\begin{figure}[t]
    \centering
    \includegraphics[width=\columnwidth]{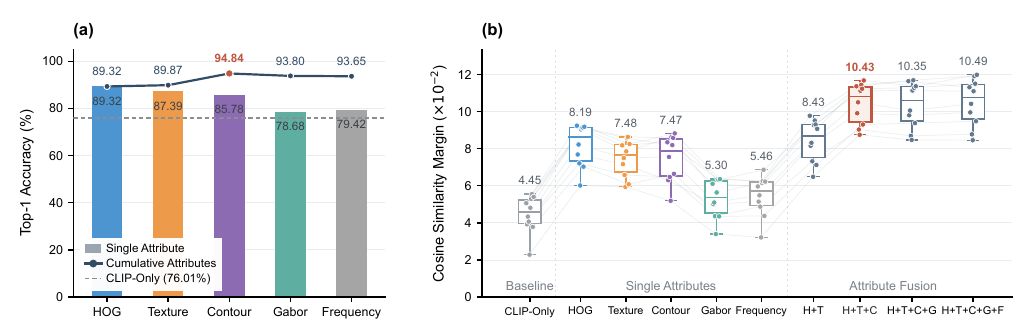}
\caption{Effect of visual attribute selection and
fusion on THINGS-EEG.
(a) Top-1 retrieval accuracy for individual and
cumulative visual attributes.
(b) Cosine similarity margins across different
attribute configurations, with subject-level
distributions and mean values.}
    \label{fig:attribute_analysis}
\end{figure}

\noindent\textbf{Visual Attribute Contribution.}
Building on the analysis in Fig.~\ref{fig:motivation1}, we examine
the effects of individual visual attributes and
their combinations on EEG-to-image retrieval.
Fig.~\ref{fig:attribute_analysis}(a) presents individual attributes (bars)
and their cumulative combinations (line), with
CLIP-only as the baseline. All individual
attributes improve Top-1 accuracy over the
baseline (76.01\%). Combining HOG, Texture,
and Contour achieves the highest Top-1 accuracy
of 94.84\%, an improvement of 18.83\%. Adding Gabor and Frequency further
reduces accuracy to 93.80\% and 93.65\%,
respectively. This observation is consistent
with their negative $\Delta R^2$ values in
Fig.~\ref{fig:motivation1}(b), suggesting limited complementary
information beyond CLIP.

To further assess retrieval discriminability,
Fig.~\ref{fig:attribute_analysis}(b) shows the cosine similarity margin
between the ground-truth image and its hardest
negative candidate. Dots represent subject-level
means, while box plots summarize the distributions
across ten subjects. Combining HOG, Texture,
and Contour increases the mean margin from
0.0445 (CLIP-only) to 0.1043, indicating
improved separation between correct and
competing images. Although adding Gabor and
Frequency produces comparable margins, it
does not improve Top-1 accuracy. These results
highlight the importance of selecting
complementary EEG-relevant visual attributes
rather than simply increasing their number.

\noindent\textbf{Effect of Spatial Correspondence.}
To assess the importance of spatial alignment
between visual attributes and regional CLIP
features, we evaluate three variants:
(1) Global Attributes, which averages attributes
across nine regions and broadcasts the result
to all regions; (2) Center-only Attributes,
which assigns the center region's attributes
to all regions; and (3) Shuffled Correspondence,
which permutes attribute vectors
within each image while preserving their content.
The permutations are fixed for each image,
and variants are trained from scratch
under identical settings.

As shown in Table~\ref{tab:spatial_ablation},
AttrEcho achieves 94.84\% Top-1 accuracy,
outperforming Global Attributes, Center-only
Attributes, and Shuffled Correspondence by
15.23\%, 10.42\%, and 14.32\%,
respectively. The substantial
performance drop under shuffled correspondence,
despite preserving all regional attribute
information, highlights the importance of
correct spatial alignment between local
visual attributes and regional CLIP features.

\begin{table}[t]
\centering
\caption{Effect of spatial correspondence in
attribute fusion on THINGS-EEG.}
\label{tab:spatial_ablation}

\footnotesize
\setlength{\tabcolsep}{3pt}
\renewcommand{\arraystretch}{1.05}

\resizebox{\columnwidth}{!}{%
\begin{tabular}{lccc}
\toprule
\textbf{Spatial Variant} &
\textbf{Top-1} &
\textbf{Top-3} &
\textbf{Top-5} \\
\midrule
Global Attributes
& $79.61 \pm 3.16$
& $93.64 \pm 1.53$
& $96.54 \pm 1.30$ \\
Center-only Attributes
& $84.42 \pm 2.81$
& $95.92 \pm 1.48$
& $98.27 \pm 0.97$ \\
Shuffled Correspondence
& $80.52 \pm 2.71$
& $94.40 \pm 2.08$
& $96.99 \pm 1.58$ \\
\midrule
\textbf{AttrEcho}
& $\mathbf{94.84 \pm 2.24}$
& $\mathbf{99.20 \pm 0.63}$
& $\mathbf{99.66 \pm 0.32}$ \\
\bottomrule
\end{tabular}%
}
\end{table}

\noindent\textbf{Effect of Aggregation Granularity.}
We compare three aggregation strategies in FTA:
Uniform Aggregation, which assigns equal weights
to all regions; Region-wise Weighting, which
learns a scalar weight per region; and
Dimension-wise Weighting, which learns separate
regional weights for each feature dimension. As shown in Table~\ref{tab:fta_ablation}, Region-wise Weighting improves Top-1 accuracy from 73.87\% to 80.30\% over Uniform Aggregation. Dimension-wise Weighting further increases
Top-1 accuracy to 94.84\%, outperforming
Region-wise Weighting by 14.54\%.
Similar improvements are observed in Top-3
and Top-5 accuracy. These results indicate that dimension-wise
weighting better captures the varying
contributions of spatial regions across
feature dimensions than region-wise weighting.
\begin{table}[t]
\centering
\caption{Ablation study of feature aggregation
strategies in FTA on THINGS-EEG.}
\label{tab:fta_ablation}
\footnotesize
\setlength{\tabcolsep}{4pt}
\renewcommand{\arraystretch}{1.3}
\resizebox{\columnwidth}{!}{%
\begin{tabular}{lccc}
\toprule
\textbf{Aggregation Strategy} &
\textbf{Top-1} &
\textbf{Top-3} &
\textbf{Top-5} \\
\midrule
Uniform Aggregation
& $73.87 \pm 3.69$
& $91.48 \pm 3.24$
& $95.20 \pm 2.38$ \\
Region-wise Weighting
& $80.30 \pm 4.22$
& $94.76 \pm 1.98$
& $97.42 \pm 1.40$ \\
\midrule
\textbf{AttrEcho}
& $\mathbf{94.84 \pm 2.24}$
& $\mathbf{99.20 \pm 0.63}$
& $\mathbf{99.66 \pm 0.32}$  \\
\bottomrule
\end{tabular}%
}
\end{table}

\subsection{Qualitative Visualization}
\begin{figure}[t]
    \centering
    \includegraphics[width=\columnwidth]
    {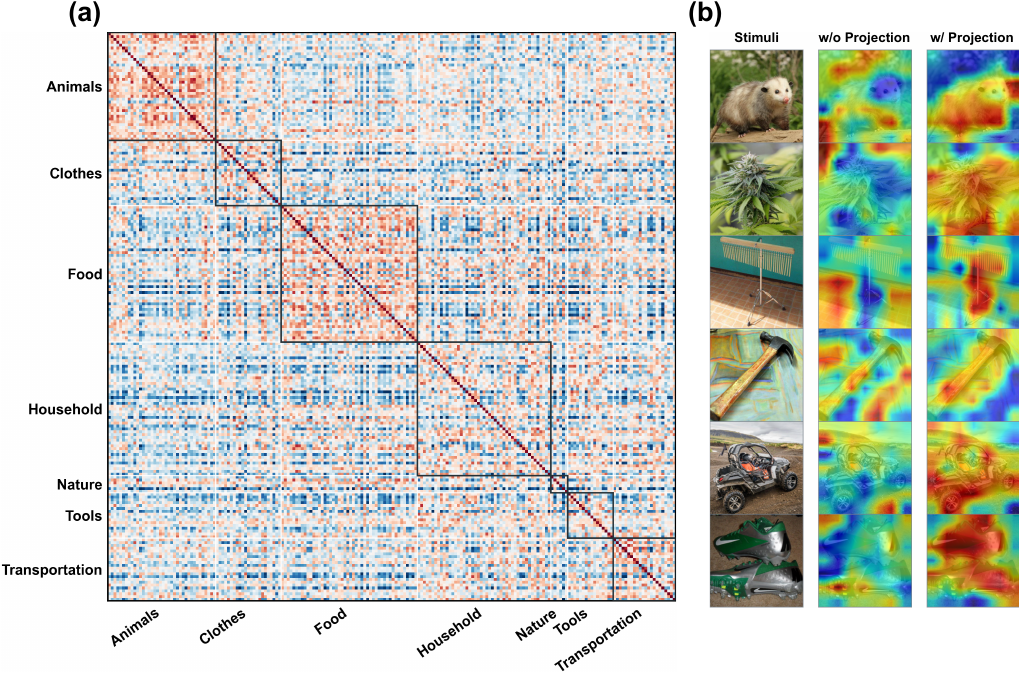}
    \caption{Qualitative visualization of visual
representations.
    (a) Category-grouped EEG--image similarity matrix.
    (b) EEG--visual
similarity maps with and without AttrEcho-trained projection.}
    \label{fig:qualitative}
\end{figure}

\noindent\textbf{Representation Visualization.}
As shown in Fig.~\ref{fig:qualitative}(a), we visualize the similarity matrix of 200 test concepts grouped into seven semantic categories. The matrix exhibits a clear diagonal pattern and noticeable within-category similarity, particularly for animals and food, indicating that the learned representations preserve category-related semantic structure. Fig.~\ref{fig:qualitative}(b) visualizes the EEG--visual
similarities of frozen CLIP features before
and after applying the image projection
learned by AttrEcho. The projected features
exhibit different spatial similarity patterns,
with high-response regions more closely
overlapping the main objects in several
examples. These observations illustrate
how the learned projection reshapes the
spatial correspondence between CLIP
features and EEG representations.

\begin{figure}[t]
    \centering
    \includegraphics[width=\columnwidth]
    {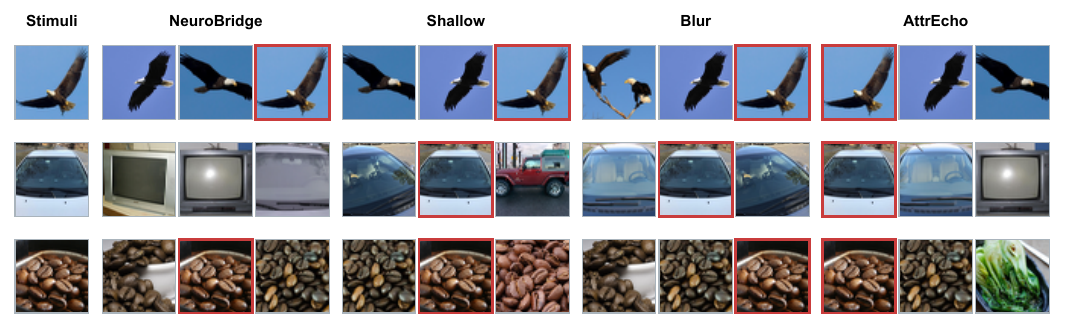}
    \caption{Top-3 retrieval comparisons on THINGS-EEG under the 1600-way setting. Red boxes indicate the ground-truth target.}  
    \label{fig:large_gallery_qualitative}
\end{figure}

\noindent\textbf{Large-Gallery Retrieval Visualization.}
Fig.~\ref{fig:large_gallery_qualitative}
presents Top-3 retrieval results of
different methods under the 1600-way setting.
In the eagle, windshield, and coffee bean
examples, competing methods often retrieve
visually similar instances but fail to rank
the exact viewed image first. In contrast,
AttrEcho ranks the ground-truth image first
in all three cases, illustrating its ability
to distinguish visually similar instances
under same-concept interference.

\begin{figure}[t]
    \centering
    \includegraphics[width=\columnwidth]
    {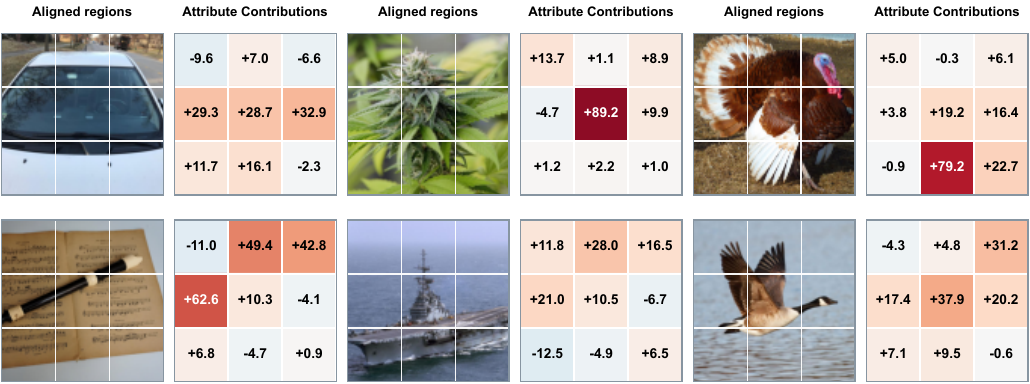}
\caption{Regional attribute contribution
visualization. Heatmaps show changes in
retrieval margin after masking regional
visual attributes. Values are reported
in units of $10^{-3}$.}
\label{fig:regional_contribution}
\end{figure}

\noindent\textbf{Regional Attribute Contribution.}
To examine the role of spatially aligned visual attributes, we conduct region-wise interventions, as shown in Fig.~\ref{fig:regional_contribution}. For each region, HOG, Texture, and Contour features are replaced with their standardized mean values, while other regions remain unchanged. Regional contributions are quantified by the reduction in retrieval margin between the ground-truth image and its hardest negative candidate after masking, where positive values indicate beneficial contributions to target discrimination. The heatmaps reveal spatial contribution patterns across images. Notably, the turkey image exhibits a strong positive contribution from the lower-central region, overlapping its body and wing. These observations complement the spatial correspondence ablation by illustrating how different regions contribute to target discrimination.

\begin{figure}[t]
    \centering
    \includegraphics[width=\columnwidth]
    {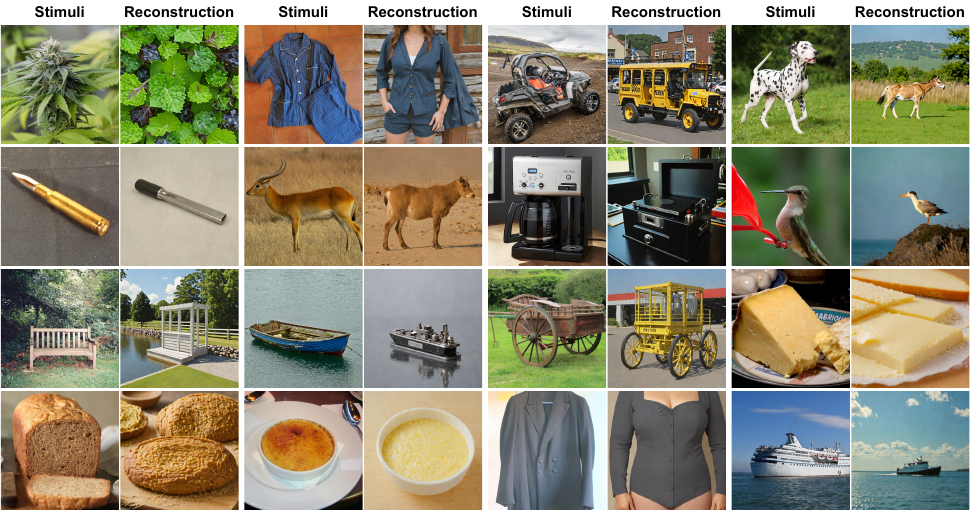}
\caption{Each pair presents the original visual stimulus (left) and the corresponding EEG-conditioned reconstructed image (right).}
\label{fig:reconstructed}
\end{figure}

\noindent\textbf{EEG-to-Image Reconstruction.}
To further evaluate the visual information captured by AttrEcho, we adopt the diffusion-based reconstruction framework of ATM \cite{li2024visual}. Specifically, EEG representations extracted by AttrEcho are used to guide image generation through a diffusion prior and SDXL with IP-Adapter. As shown in Fig.~\ref{fig:reconstructed}, the reconstructed images preserve semantic and visual characteristics of the stimuli, although discrepancies in fine-grained details remain. These results demonstrate the potential of AttrEcho representations for visual reconstruction. 

\section{Conclusion}
In this paper, we investigate how explicit visual attributes complement pretrained visual representations for EEG-to-image retrieval. We propose AttrEcho, a region-aligned attribute fusion framework that integrates complementary visual cues with regional CLIP features through gated residual fusion and feature aggregation. Extensive experiments demonstrate its effectiveness in intra-subject and inter-subject retrieval, including large galleries with same-concept distractors. Ablation studies and regional interventions further highlight the importance of EEG-relevant attribute selection, spatial correspondence, and feature aggregation. Overall, our findings suggest that effective EEG-to-image alignment benefits not simply from incorporating more visual attributes, but from selecting complementary cues and preserving their spatial correspondence.

\section*{LLM Usage Statement}
We used OpenAI’s ChatGPT to assist with language editing, LaTeX formatting, visualization refinement, and limited code development. The authors reviewed and validated all AI-assisted outputs, independently verified the analyses, results, figures, and scientific claims, and take full responsibility for the final manuscript.

{
    \small
    \bibliographystyle{ieeenat_fullname}
    \bibliography{main}
}


\end{document}